# An United Image Force for Deformable Models and Direct Transforming Geometric Active Contorus to Snakes by Level Sets

Hongyu Lu
Department of Computer Science
Beijing Institute of Petrochemical Technology
Beijing, China, 102617
LUHongyu@bipt.edu.cn

Yutian Wang
The First Research Institute of
Ministry of Public Security of P.R.C
Beijing, 100871, China
ytwang99@126.com

Shanglian Bao
School of Physics
Peking University
Beijing, 100048, China
Bao@pku.edu.cn

*Abstract*: **A uniform distribution of the image force field around the object fasts the convergence speed of the segmentation process. However, to achieve this aim, it causes the force constructed from the heat diffusion model unable to indicate the object boundaries accurately. The image force based on electrostatic field model can perform an exact shape recovery. First, this study introduces a fusion scheme of these two image forces, which is capable of extracting the object boundary with high precision and fast speed. Until now, there is no satisfied analysis about the relationship between Snakes and Geometric Active Contours (GAC). The second contribution of this study addresses that the GAC model can be deduced directly from Snakes model. It proves that each term in GAC and Snakes is correspondent and has similar function. However, the two models are expressed using different mathematics. Further, since losing the ability of rotating the contour, adoption of level sets can limits the usage of GAC in some circumstances.**

## Ⅰ. Introduction

Active contours or deformable models have been widely applied in pattern recognition for image analysis, especially in segmentation. Active contours are classified into two groups: Parametric Active Contours (PAC or Snakes), and Geometric Active Contours (GAC or Geodesic Active Contours). Both of them can be achieved through edge based approach [1] [2] or region based approach [3] [4]. Each algorithm has its own advantage in processing different kinds of images. In this study, only the deformable models based edge map are considered.

The edge map reflects the crucial features of an image. It can be utilized to generate the image force for attracting the Snakes or GAC contours to the desired object boundary. The accurate recovery of the boundary and the fast evolution of the contours are important to the segmentation process. The image force can be constructed from the electrostatic field model or the heat diffusion model. Most of the current models follow these two ways. The drawbacks of both models are studied deeply in the following section. Then a combined approach is developed to meet the requirements for accuracy and speed.

The expressions of the Snakes model and GAC model [5] seem quite different. There is a term containing the curvature in GAC. Further, Snakes is presented through differential equations and evolving in sub-pixels resolution. GAC is built on the basis of level sets theory. A strict procedure of formula derivation from Snakes to GAC is introduced to investigate the links between two models. It addresses that the two models are almost the same in principles indeed. The specialty of GAC is that solving equations through level sets approach makes the contour move only in the normal direction.

## II. Analysis of the Image force by using Snakes

According to elastic theory, Snakes can be represented by

$$\frac{\partial C(t)}{\partial t} = \alpha \frac{\partial^2 C(t)}{\partial s^2} + \beta \vec{N}(x,y) + \gamma \vec{F}_{img}(x,y) \qquad (1)$$

Where $\vec{F}_{img} = (f_x, f_y)$ represents the image vector force and $\vec{N} = (n_x, n_y)$ represents unit normal. Both of them are expressed in $x, y$ directions. The first term on the right side of (1) is an elastic force, which works like that in a spring. The second term is the balloon force, and the third one is the image force. Since it has $C(t) = (x(t), y(t))$, the equation (1) can be rewritten as

$$\frac{\partial x}{\partial t} = \alpha \frac{\partial^2 x}{\partial s^2} + \beta n_x(x,y) + \gamma f_{img\_x}(x,y) \quad (2a)$$

$$\frac{\partial y}{\partial t} = \alpha \frac{\partial^2 y}{\partial s^2} + \beta n_y(x,y) + \gamma f_{img\_y}(x,y) \quad (2b)$$

It can be seen that the image force play a key role in Snakes. This is also true in GAC method. As described in [6], the potential filed of a time-invariant general force can take the format as

$$P(x,y) = \sum_{y_i}\left[\sum_{x_i} g_{x_i,y_i} \cdot f(r)\right] \quad (3)$$

Where $g_{x_i,y_i}$ represents the gray level of the point at $(x_i, y_i)$ of the edge map on the image plane. It has: $r = \sqrt{(x-x_i)^2 + (y-y_i)^2 + h^2}$, and $f(r)$ is the transfer function responsible for creation of the image force. As the experiments point out, the format of the equation (3) based on electrostatic field exhibits the best segmentation performance in most cases. It can be outlined as

$$P = \frac{1}{k \cdot r^\lambda} \quad (4)$$

Generally, $\lambda$ adopts the value such as 1, 1/2, 1/3, 1/4. The image force can be described as

$$F_{img\_x}(x,y) = \frac{\partial P(x,y)}{\partial x} \quad (5a)$$

$$F_{img\_y}(x,y) = \frac{\partial P(x,y)}{\partial y} \quad (5b)$$

The problem of this image force field is that strengths of the force fall down quickly when it is far away from the object edges. The GVF model [7] overcomes this problem by using the theory of the optical flow. This time-varying method gives a more even distribution of the force field around the object boundary. It means the intensity of the force does not change too much in the distance of 10-20 pixels away from the edge map. This complicated force model brings the problems to the shape having T-junction. The troubles are: 1) the segmentation accuracy is not good at the corners as shown in Fig. 1 (a). 2) the image force has the same downward direction in the tubular area as pointed in Fig. 1 (b). The strength of the forces is very similar. If the initial contour indicated by the green circle in Fig. 1 (a)., is inside the "T" Shape and is placed at the bottom, it makes the curve difficult to move upward to the location of T-junction even if a balloon force drives it. 3) the computational time is very long.

The image force constructed from electrostatic field [8] does not have the three problems as GVF, as shown in Fig. 2. However, the force field is not smooth enough to support the fast speed of evolution of the active contours. The heat or thermal diffusion equation reflects a dynamic system, which is able to generate an even force field around the object. It is given by

$$\frac{\partial T}{\partial t} = \frac{\partial^2 T}{\partial x^2} + \frac{\partial^2 T}{\partial y^2} \quad (6)$$

Treating the gray level of each pixel as the temperature value, it can produce a potential field of the temperature through thermal diffusion in a time period. The longer the diffusing time, the smoother the image force field constructed, and the worse segmentation accuracy got. The image force field and segmentation result in Fig. 3 shows these points.

III. The United image force

To avoid the error in extracting the boundaries, and to accelerate the speed of segmentation process, it is applicable to put the force model based on electric field and thermal diffusion together. This idea can be expressed by

$$\vec{F}_{img} = \gamma_e \vec{F}_{Elect} + \gamma_h \vec{F}_{Heat} \quad (7)$$

It is reasonable that the heat force contributes more to $\vec{F}_{img}$ at the far position and contribute less to $\vec{F}_{img}$ at the near position from the object boundary. The formula (7) is reformatted to

$$\vec{F}_{img} = \gamma_e \vec{F}_{Elect} + [k(t) \cdot \gamma_h]\vec{F}_{Heat} \quad (8)$$

Where value of $k(t)$ reduces according to time or the iteration number, such as taking the format: $k(t) = 1/(1 + \log(Iterations))$. An artificial circle is processed using the Snakes equations (2) and (3), and the image force model (4), (6) and (8) separately. Three segmentation results are show in Fig. 4 for comparing three different methods above. There is 2-pixels segmentation error in the result coming from the heat diffusion model in Fig. 4 (b). It means that the peak of the potential field moves inside to the circle several pixels, which is illustrated clearly in the Fig. 5. The longer the diffusion times, the bigger the error appears. This phenomenon is similar to what is described in [8]. It is also indicated on the page 23, chapter 2 in [9] and [10], which needs further study. The electrostatic force model gives some explanations for operators like Gaussian. Another possible reason is that the object has closed boundary, which likes a circle. The convergence time of three image force models for segmenting the "T" shape image is given in Table 1, which are tested on the Intel core i5 2.6G CPU. It can be seen that the united image force grantees speed. The initializations of the contour are same in Fig. 1, 2 and 3. The main parameters are set to the best value that can segment the "T" shape with the best result in the fast time.

Ⅳ. Construction of the GAC equation form Snakes model

According to Frenet-Serret formulas, since the unit tangent $\bar{T} = \partial C/\partial s$, it can get $\partial^2 C/\partial s^2 = \partial \bar{T}/\partial s = \kappa \bar{N}$, where $\kappa$ is the curvature. Apparently, the acceleration vector in physics is got in the same way. The equation (1), becomes as

$$\frac{\partial C(t)}{\partial t} = \alpha \kappa \bar{N} + \beta \bar{N} + \gamma \bar{F}_{img} \quad (9)$$

Holding by level sets theory, it has

$$C(t) = \{(x, y) \mid \phi(x, y, t) = 0\} \quad (10)$$

It is known that $\phi(C(t),t) = 0$ so that the partial derivatives of demanding $\phi(C(t),t)$ on the $t$ is

$$\frac{\partial \phi}{\partial C} \bullet \frac{\partial C}{\partial t} + \frac{\partial \phi}{\partial t} = 0 \quad (11)$$

Equation (11) can be transferred to

$$\frac{\partial \phi}{\partial t} = -\frac{\partial \phi}{\partial C} \bullet \frac{\partial C}{\partial t} = -\nabla \phi \bullet \frac{\partial C}{\partial t} \quad (12)$$

Where $\nabla \phi = \frac{\partial \phi}{\partial C} = \left(\frac{\partial \phi}{\partial x}, \frac{\partial \phi}{\partial y}\right) = (\phi_x, \phi_y)$

Therefore, it can get

$$\frac{\partial \phi}{\partial t} = -\nabla \phi \bullet \frac{\partial C}{\partial t} = -\nabla \phi \bullet \left[\alpha \kappa \bar{N} + \beta \bar{N} + \gamma \bar{F}_{img}\right] \quad (13)$$

Since $\bar{N} = -\frac{\nabla \phi}{|\nabla \phi|}$, it can get

$$\frac{\partial \phi}{\partial t} = \alpha \kappa \left(\nabla \phi \bullet \frac{\nabla \phi}{|\nabla \phi|}\right) + \beta \left(\nabla \phi \bullet \frac{\nabla \phi}{|\nabla \phi|}\right) - \gamma \bar{F}_{img} \bullet \nabla \phi \quad (14)$$

Since it has $\nabla \phi \bullet \nabla \phi = |\nabla \phi|^2$, then the GAC equation can be expressed as

$$\frac{\partial \phi}{\partial t} = \alpha \kappa |\nabla \phi| + \beta |\nabla \phi| - \gamma \bar{F}_{img} \bullet \nabla \phi \quad (15)$$

This is the formula used in [5]. It means that the each term on the right side of equation (15) has similar functions as that in the Snakes equation (1). Hence, it addresses that Snakes and GAC comes from a very similar model. It should be noticed that the equation (15) is correct only if $\phi = 0$. Since the usage of extension velocities or force brings more errors and is difficult to calculate, this projection is not required if narrow band method is applied.

To fast the calculation and improve the segmentation accuracy in low order, the third term should be replaced the sign of the $\bar{F}_{img} \bullet \nabla \phi$ [6]. Hence, the equation (15) be reformatted as,

$$\frac{\partial \phi}{\partial t} = \frac{\partial \phi}{\partial t} = \alpha \kappa |\nabla \phi| + \beta |\nabla \phi| - \gamma \text{sign}(\bar{F}_{img} \bullet \nabla \phi) \quad (16)$$

The equation (16) gives the better result than most high-order schemes. The numerical solution method of $\bar{F}_{img} \bullet \nabla \phi$ by narrow band method might be the reason.

It should be noticed that Xu established a relationship between PAC and GAC concisely in paper [13]. However, his deduction was built on the assumption that GAC must adopt the force components normal to the contour. The aim of the assumption is to drive the contour to the destination in the shortest path on the basis of the Euler-Lagrange equation. In the procedure above, it is the equations (11) and (12) based on the level sets require the forces components normal to contour, not as what said in [13]. Mihalef also discovered this problem in paper [14].

Actually, $\vec{T}$ term can not be ignored in most cases because of its power to rotate a contour. It is also capable of changing geometry of shape [15]. Level sets using only one variable $\phi$ to represent the curve in 2-D. Due to this reason, it is easy to see from the definition that $\phi$ lose the capability to rotate the contour during the curve evolution. However, the Snakes PDE equations can make this rotation easily. Hence, it is should be careful to apply the level sets in image analysis, patter recognition, computer vision, etc.

Ⅴ. Experiments and Results

Four medical images are tested. The Fig. 6 shows extraction of a brain tumor. It is processed by Snakes model with adopting equations (2) and (8). The shape recovery of the eight bacteria in Fig. 7 (a) utilizes a kind of Snakes with the ability to extract multi-objects [12]. Its algorithm depends on the same equations for segmentation of brain tumor. The result is shown in Fig. 7 (b). A 2-D liver image is successfully recovered in Fig. 8. The proposed GAC algorithm can be extend to a 3-D scheme. Fig. 9 shows the extraction of the bladder from a CT image datasets. The experiments demonstrate that our approaches can minimize the errors in the shape extraction.

Ⅵ. Conclusion

The shortcomings of the current image force models are studied to seek an effective infrastructure for generating the reliable image force. An approach based on the dynamic heat diffusion is designed to complement the time-invariant electrostatic field model. It can create a satisfied image force field to improve the segmentation accuracy and the convergence speed for the active contours. Further, the mathematic deduction demonstrates that the GAC equation can be derived directly from original expressions of the Snakes by level sets. It proves that the terms in Snakes and GAC has almost the same meaning, but are represented in different mathematics approaches. The adoption of level sets restrains the utilization of the GAC in some fields. Therefore, the techniques using level sets to process image such as the C-V model demands investigations carefully in some cases.

## References

[1] M. Kass, A.W., D. Terzopoulos, "Snakes: Active contour models", International Journal of Computer Vision, 1988, vol. 1, pp. 321-331.

[2] V. Caselles, R. Kimmel, G. Sapiro, "Geodesic Active Contours", International Journal of Computer Vision, 1997, vol. 22, pp. 61-79.

[3] T. McInerney, D. Terzopoulos, "T-Snakes: topology adaptive snakes", Medical Image Analysis, 2000, vol 4, pp. 73-91.

[4] T. F. Chan, L. A. Vese, "Active Contour without Edges", IEEE Transactions on Image Processing, 2001, vol. 10, pp. 266-277.

[5] R. Malladi and J.A. Sethian. An O(n logn) Algorithm for Shape Modeling. Proceedings of the National Academy of Sciences, 1996, Vol. 93, pp. 9389–9392.

[6] H. Y. Lu, C. J. Duan, S. L. Bao. "Accurate Segmentation of Image with Large Gaps by Geometric Active Contours," in Proceedings of IEEE 9th International Conference on Signal Processing, Beijing, China, Oct. 2008, pp. 1235-1375.

[7] C. Xu, and J. L.Prince, "Snakes, Shapes, and Gradient Vector Flow", IEEE Trans. on Image Processing, 1998, vol. 7, no. 3, pp. 359-369.

[8] H.Y. Lu, S.H. Bao, D.L. Zu, "Application of electrostatic field in snakes for medical image segmentation", in Proceedings of IEEE 7th International Conference on Signal Processing, Beijing, China, Sept., 2004, vol. 1, pp.793-796.

[9] S.D Ma, Z.Y. Zhang, "Computer Vision: : The Fundamentals of computational theory and algorithm"(in Chinese), Chinese Science Press, Beijing, 1998.

[10] H. Y. Lu, S. L. Bao. "An Extended Image Force Model of Snakes for Medical Image Segmentation and Smoothing," in Proceedings of IEEE 8th International Conference on Signal Processing, Guilin, China, Nov. 2006,   pp. 1372-1375.

[12] H.Y. Lu, S.L Bao. "Physical Modeling Techniques in Active Contours for Image Segmentation".
http://arxiv.org/ftp/arxiv/papers/0906/0906.4036.pdf, June, 2009.

[13] C. Xu, A. Yezzi, and J. L. Prince. "On the relationship between parametric and geometric active contours". Conference Record of the Thirty-Fourth Asilomar Conference on Signals, Systems and Computers, vol. 1, pp. 483 – 489, 2000.

[14] V. Mihalef, M. Sussman, D. Metaxas. "The Marker Level Set method: a new approach to computing accurate interfacial, dynamics", www.research.rutgers.edu/~mihalef/mls_jcp.pdf

[15] http://www.math.uni-hamburg.de/projekte/shape/

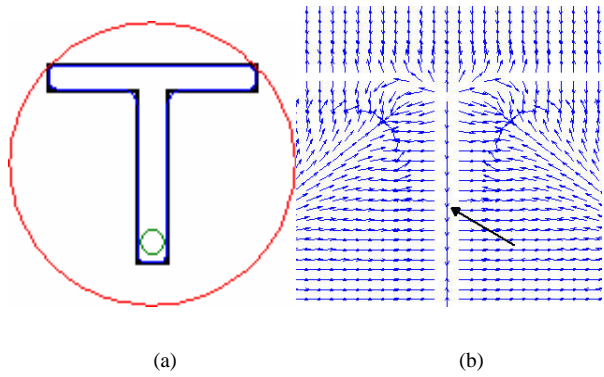

Figure 1. Segmentation of the artificial "T" (Black) with GVF image force

(a) Initialization of the Snakes contour (Red) and the segment result (Blue)

(b) Force field created by GVF at the T-junction.

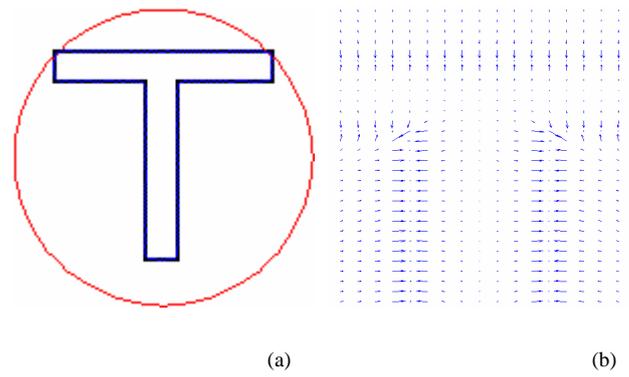

Figure 2. Segmentation of the artificial "T" (Black) with image force based on electrostatic field

(a) Initialization of Snakes contour (Red) and the segment result (Blue)

(b) Force field created by electrostatic field model at the T-junction

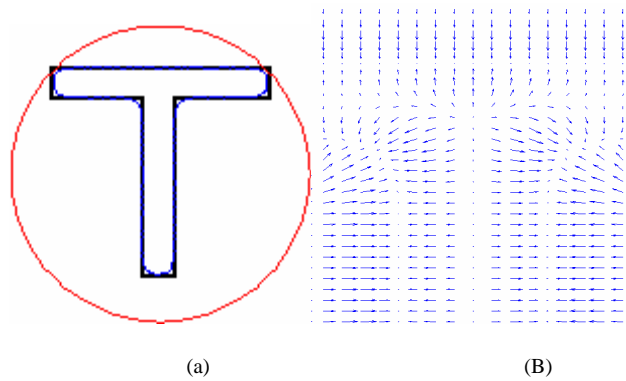

Figure 3. Segmentation of the artificial "T" (Black) by image force based on the heat diffusion

(a) Initialization of the Snakes contour (Red) and the segment result (Blue)   (b) Force field created by heat diffusion model at the T-junction

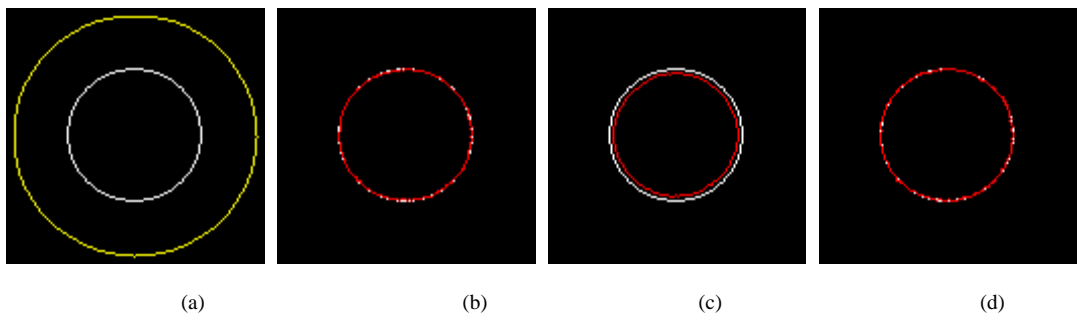

Figure 4.  Segmentation of an artificial circle (White) by Snakes using different image forces

(a) Initialization contour of Snakes (Yellow) ( b) Result of electrostatic field   (c) Result of heat diffusion   (d) Result of united force (Red)

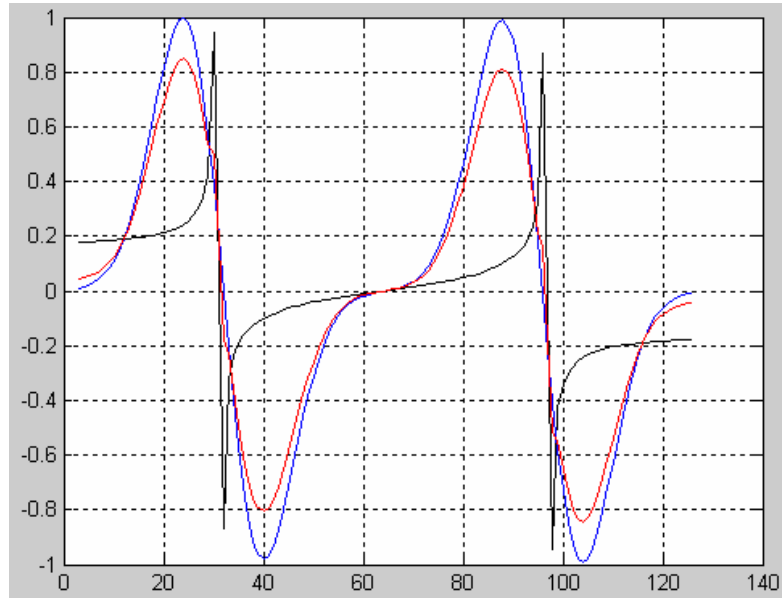

Figure 5. The potential created by the three image force models along the x-axis of the circle

Electrostatic field(Black), Heat diffusion (Blue), Combination of electrostatic field and heat diffusion(Red)

Table 1. Comparison of the Snakes converge speed using different image forces of "T" Shape

| Method | Electrostatic field | Heat diffusion | United method |
|---|---|---|---|
| Computing Time | 0.226s | 0.152s | 0.173s |

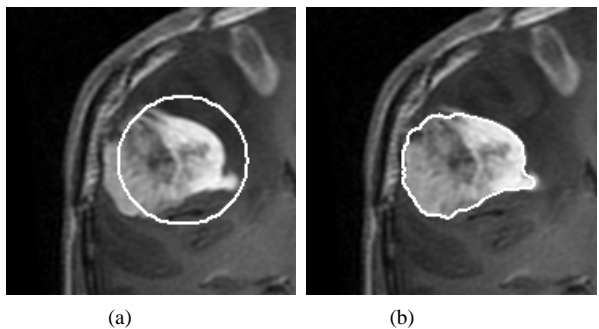    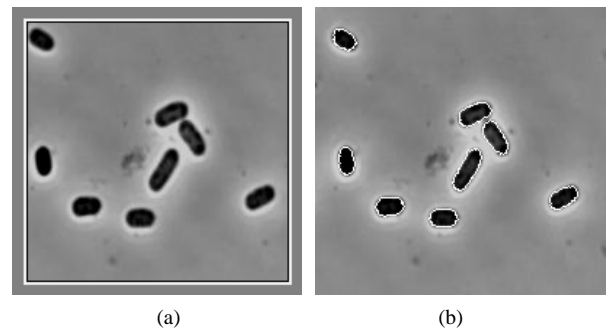

(a)        (b)                    (a)        (b)

Figure 6. The segmentation of a brain tumor          Figure 7. The segmentation of a brain tumor of eight bacteria

(a) Original Image and Initialization of Snakes Contours (White)   (a) Original Image and Initialization of Snakes Contours (white rectangle)

(b) The Segmentation result                    (b) The Segmentation result

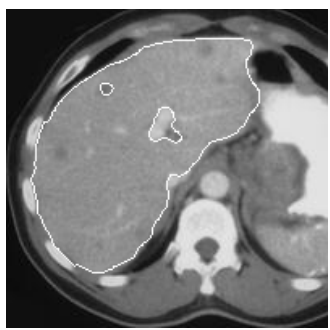    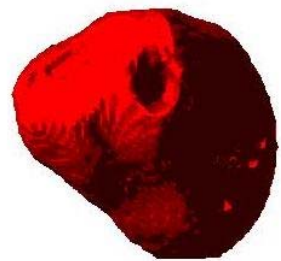

Figure 8. Segmentation result of the liver by GAC     Figure 9. The 3-D segmentation result a CT bladder by GAC